\pdfoutput=1
\documentclass[letterpaper]{article}
\usepackage[preprint]{aaai2027}
\usepackage[hyphens]{url}
\usepackage{graphicx}
\usepackage{natbib}
\usepackage{caption}
\usepackage{subcaption}
\usepackage{booktabs}
\usepackage{multirow}
\usepackage{amsmath}
\usepackage{amssymb}
\usepackage{listings}
\usepackage{xspace}
\newcommand{\projecturl}{%
  \leavevmode
  \pdfstartlink attr{/Border [0 0 0]} user{%
    /Subtype /Link
    /A << /Type /Action /S /URI
    /URI (https://github.com/ds-kiel/ProgResViT) >>}%
  \url{https://github.com/ds-kiel/ProgResViT}%
  \pdfendlink
}

\definecolor{progressblue}{HTML}{4285F4}
\definecolor{progressgreen}{HTML}{22C55E}
\NewDocumentCommand{\name}{s}{%
  \IfBooleanTF{#1}{%
    \textcolor{progressblue}{Prog}\textcolor{progressgreen}{Res}ViT%
  }{ProgResViT}%
  \xspace
}

\newcommand{\appref}[1]{Appendix~\ref{#1}}
\newcommand{\Appref}[1]{Appendix~\ref{#1}}

\title{\name: Progressive Resolution and Width for Adaptive Vision Transformers}
\author{
  Ali Hojjat\textsuperscript{\rm 1,\rm 2},
  Janek Haberer\textsuperscript{\rm 1},
  Olaf Landsiedel\textsuperscript{\rm 2,\rm 1,\rm 3}
}
\affiliations{
  \textsuperscript{\rm 1}Kiel University, Germany\\
  \textsuperscript{\rm 2}Hamburg University of Technology (TUHH), Germany\\
  \textsuperscript{\rm 3}UNU-INWEH, Germany\\
  ali.hojjat@tuhh.de, janek.haberer@cs.uni-kiel.de, olaf.landsiedel@tuhh.de\\[-0.15em]
  {\scriptsize\color{blue}\projecturl}
}

\begin{document}
\maketitle

\begin{abstract}
Vision Transformers (ViTs) typically process every image using a fixed input resolution and model width, even though many images can be classified with substantially less computation. We introduce \name, an input adaptive ViT that performs inference progressively across multiple rounds. The first round processes a low-resolution image with a narrow subnetwork. Inference terminates when the prediction is sufficiently confident; otherwise, the model reuses the representations produced in the current round and proceeds with a higher-resolution input and a wider subnetwork to refine its prediction. As all rounds share a single backbone, we propose \textbf{P}rogress-Conditioned \textbf{S}oft \textbf{G}ating (PSG), which conditions token fusion and layer outputs on the current round, block, and input resolution. On image classification, applying \name to DeiT yields better accuracy--compute trade-offs than adaptive-width, adaptive-depth, and dynamic-token baselines.
With knowledge distillation, a DeiT-based \name achieves 84.9\% top-1 accuracy, slightly exceeding the reported DeiT-III-S accuracy under a comparable evaluation setting.
We show that the same design also provides favorable accuracy--compute trade-offs for self-supervised DINO representations and downstream semantic segmentation. Code is available in supplementary material.
\end{abstract}

\section{Introduction}

\paragraph{Motivation.} Vision Transformers (ViTs) process every image with the same input resolution, token grid, depth, and model width \cite{dosovitskiy2021an}. This fixed-cost design ignores two important sources of variation: images differ in difficulty, and correct predictions require different amounts of spatial detail. A clear, centered object can be recognized from a coarse view and a small model, whereas cluttered or fine-grained examples require both a denser token grid and greater representational capacity.

\paragraph{Prior work.}
Adaptive ViTs address this mismatch by varying individual computational axes. Early-exit methods adapt the executed depth
\cite{bakhtiarnia2021multi,xu2023lgvit}; token-adaptive methods prune, merge, or selectively process spatial tokens
\cite{rao2021dynamicvit,meng2022adavit,yin2022vit}; and adaptive-width Transformers expose subnetworks from a single parameter set
\cite{devvrit2024matformer}.
ThinkingViT is the closest predecessor to our setting: it repeatedly executes progressively wider subnetworks and routes images according to prediction uncertainty
\cite{hojjat2026thinkingvit}.
However, its rounds retain a fixed input resolution, and it does not explicitly condition the repeatedly executed shared blocks on the current stage. In general, existing adaptive ViTs typically adjust only one computational axis at a time or rely on separately optimized operating points, leaving open how a single shared model can jointly adapt spatial resolution and model capacity while progressively refining representations.

\begin{figure*}[!t]
  \centering
  \includegraphics[width=\textwidth]{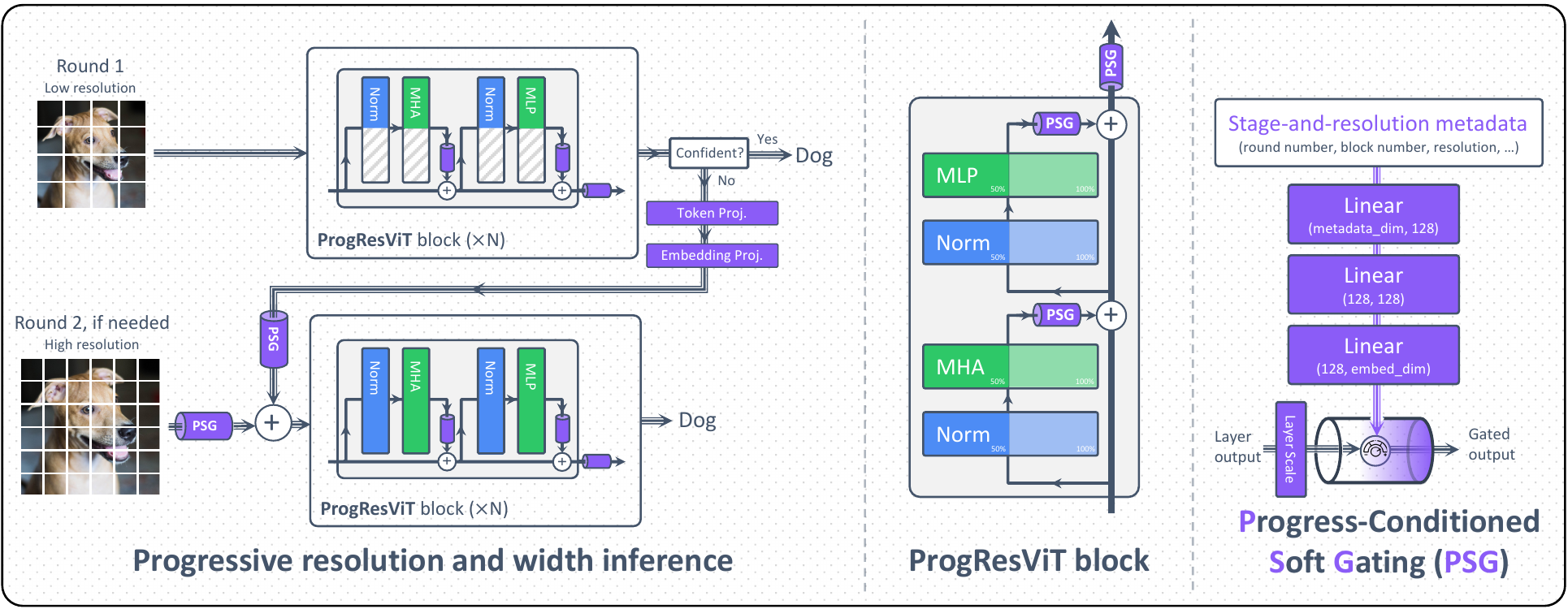}
  \caption{Overview of \name: Round 1 processes a low-resolution image using a narrow subnetwork with entropy-based routing. The second round uses a higher-resolution version of the input to generate new embeddings. The round 1 tokens are projected to match the dimensions of the new embeddings and then merged with them. The resulting combined embeddings are processed by the full subnetwork. As the same backbone is reused across rounds, its shared blocks must process representations with different token grids, spatial resolutions, and active widths, despite receiving no explicit indication of the current stage. To address this mismatch, \name introduces \textbf{P}rogress-Conditioned \textbf{S}oft \textbf{G}ating (PSG), which uses round, block, and resolution metadata to condition token fusion and layer outputs on the current stage.}
  \label{fig:overview}
\end{figure*}

\paragraph{\name.}
We introduce \name, a confidence-routed ViT that progressively increases both image resolution and active model width. The model follows an ordered sequence of resolution--width operating points. The first round processes a low-resolution image with a narrow subnetwork. Inference terminates when the prediction is sufficiently confident; otherwise, a cross-resolution token projector spatially aligns the patch tokens, expands their embedding dimension, and fuses them with fresh higher-resolution embeddings. The model then activates the full-width subnetwork, using more channels and attention heads and thereby allocating additional computation to harder samples. Consequently, a single trained model provides a tunable accuracy--compute trade-off by varying only the inference threshold, without requiring separate models for different operating points. Figure~\ref{fig:overview} summarizes the pipeline.

Iterative inference uses the same Transformer blocks under different operating conditions, including changes in input resolution and active model width. Naively reusing the same blocks across these settings forces each block to process representations with different spatial and channel characteristics without explicit knowledge of the current stage.
Inspired by \cite{jacobs2026raptor}, we introduce \textbf{P}rogress-Conditioned \textbf{S}oft \textbf{G}ating (PSG), a stage-conditioned modulation mechanism that uses round, block, and resolution metadata to modulate token fusion, attention and MLP residual updates, and block outputs. This conditioning enables the shared weights to specialize across progressive stages while remaining part of a single backbone.

\begin{figure*}[!t]
  \centering
  \subcaptionbox{DINO 20-NN.\label{fig:main-frontier-dino}}[.241\textwidth]{%
    \includegraphics[width=\linewidth]{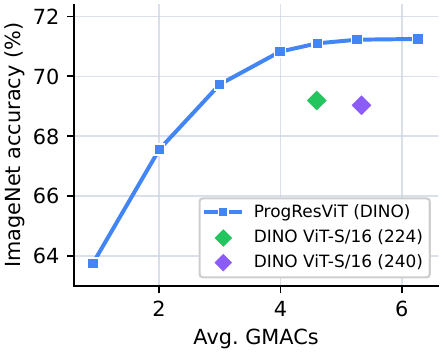}}\hfill
  \subcaptionbox{ADE20K segmentation.\label{fig:main-frontier-seg}}[.241\textwidth]{%
    \includegraphics[width=\linewidth]{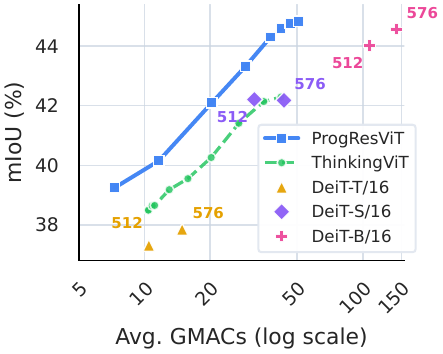}}\hfill
  \subcaptionbox{Adaptive-width baselines.\label{fig:main-frontier-imagenet}}[.241\textwidth]{%
    \includegraphics[width=\linewidth]{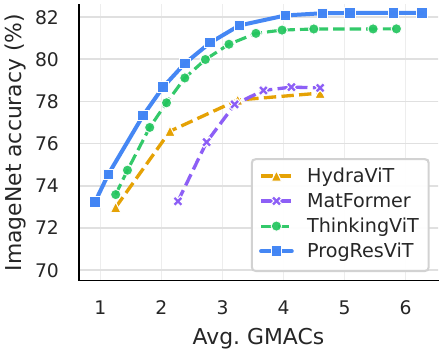}}\hfill
  \subcaptionbox{ImageNet-1K accuracy.\label{fig:main-frontier-training}}[.241\textwidth]{%
    \begin{minipage}[c]{\linewidth}
      \centering
      \small
      \setlength{\tabcolsep}{2pt}
      \renewcommand{\arraystretch}{1.3}
      \begin{tabular}{@{}lcr@{}}
        \toprule
        Model & GMACs & Top-1 \\
        \midrule
        DeiT-III-S & \multirow{2}{*}{15.5} & \multirow{2}{*}{84.80} \\
        ($384^2)$ & & \\
        \midrule
        \bfseries \name & \multirow{2}{*}{\bfseries 12.3} &
        \multirow{2}{*}{\bfseries 84.90} \\
        \bfseries ($160^2\!\rightarrow\!384^2$) & & \\
        \bottomrule
      \end{tabular}
      \par\vspace*{7mm}
    \end{minipage}}
\caption{\textbf{Summary of the main results.}
  Across (a) DINOv1 representation learning, (b) ADE20K semantic segmentation, and (c) ImageNet-1K adaptive-width inference, \name consistently outperforms the corresponding baselines at comparable compute. (d) Although built on DeiT-S, distilled \name reaches 84.90\% top-1 accuracy, slightly exceeding the reported DeiT-III-S accuracy under a comparable evaluation setting. For \name, the reported accuracy and GMACs use entropy-based routing with threshold $\tau=0.15$.}
  \label{fig:main-frontier}
\end{figure*}

\paragraph{Results Overview.}
On ImageNet-1K, \name with the $192\!\rightarrow\!240$ resolution schedule reaches 82.21\% top-1 accuracy at 6.27 GMACs under full two-round inference and retains 82.18\% at 4.47 average GMACs with confidence-based routing, reducing computation by 28.7\%. This schedule outperforms the adaptive-width, adaptive-depth, and dynamic-token baselines. For maximum accuracy, the distilled $160\!\rightarrow\!384$ resolution schedule reaches 84.90\% at 16.15 GMACs and retains 84.87\% at 11.12 average GMACs, reducing computation by 31.1\%. We further show that \name maintains frontier accuracy--compute trade-offs for DINO representation learning, semantic segmentation, and distribution shifts represented by ImageNet variants. Figure~\ref{fig:main-frontier} summarizes the main results.

\paragraph{Contributions.}

\begin{itemize}
  \item We introduce a progressive resolution--width ViT that allocates both spatial detail and model capacity according to input complexity.
  
  \item We introduce cross-resolution feature reuse, together with PSG for stage-conditioned modulation of repeatedly executed shared blocks.
  
  \item We evaluate the resulting design through classification, distribution-shift tests, self-supervised DINO representations, and semantic segmentation.
\end{itemize}

\section{Related Work}

\paragraph{Adaptive-width Transformers.}
Matryoshka Representation Learning learns useful representations at nested dimensionalities \cite{kusupati2022matryoshka}, while DynaBERT, SortedNet, MatFormer, HydraViT, and SlicingViT embed subnetworks with different capacities within a single parameter-sharing Transformer \cite{hou2020dynabert,valipour2023sortednet,devvrit2024matformer,haberer2024hydravit,zhang2024slicing}. These methods primarily support elastic deployment across selected operating points. ThinkingViT extends this idea to input-adaptive progressive inference by executing progressively wider subnetworks and applying confidence-based early termination \cite{hojjat2026thinkingvit}. \name extends this paradigm to progressive resolution and width inference.
Additionally, it transfers features across rounds by projecting previous-round representations across both the token grid and the embedding dimension. Moreover, motivated by the depth-conditioned residual scaling of Raptor \cite{jacobs2026raptor}, \name introduces \textbf{P}rogress-Conditioned \textbf{S}oft \textbf{G}ating (PSG) to explicitly condition token fusion and repeatedly executed shared blocks on the current round, block, and resolution.

\paragraph{Resolution-flexible Transformers.}
FlexiViT and related multi-resolution methods support varying patch sizes, resolutions, or aspect ratios, but do not adapt resolution to prediction difficulty \cite{beyer2023flexivit,fan2024vitar,tian2023resformer,dehghani2023navit}.
DVT and related adaptive multi-resolution methods allocate additional spatial computation to uncertain inputs \cite{wang2021not,yang2020ranet,guidez2026ravit}.
CF-ViT refines informative patches while retaining the remaining regions at a coarse scale, and LF-ViT localizes a class-discriminative region from a low-resolution image before processing that region at a higher resolution \cite{chen2023cfvit,hu2024lfvit}.
MSViT instead selects a coarse or fine token scale for every image region and preserves full-image coverage \cite{havtorn2023msvit}. Unlike these methods, \name jointly increases full-image resolution and active width within one shared backbone.

\paragraph{Early exit and token-adaptive inference.}
LGViT and other early-exit networks adapt depth using intermediate classifiers \cite{xu2023lgvit,teerapittayanon2016branchynet,huang2017multi,bakhtiarnia2021multi}. Token-adaptive methods instead modify which spatial tokens are processed. DynamicViT progressively prunes tokens according to input-dependent importance scores, A-ViT adaptively halts computation for individual tokens, and AdaViT jointly selects patches, attention heads, and Transformer blocks on a per-input basis \cite{rao2021dynamicvit,yin2022vit,meng2022adavit}. Unlike these methods, \name preserves the complete token grid at each round while adapting computation along two axes. It changes the token count by varying the input resolution and adjusts the computation per token by varying the Transformer width.

\begin{figure*}[!t]
  \centering
  \subcaptionbox{Resolution schedules.\label{fig:schedules}}[.31\textwidth]{%
    \includegraphics[width=\linewidth]{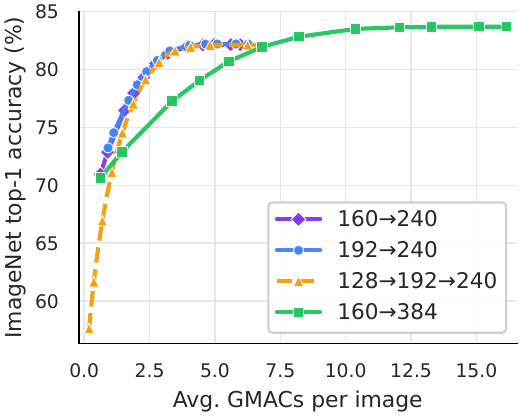}}
  \subcaptionbox{\textbf{P}rogress-Conditioned \textbf{S}oft \textbf{G}ating (PSG) multipliers.\label{fig:psg-profile-values}}[.68\textwidth]{%
    \includegraphics[width=\linewidth, trim={0.33cm 0 0 0}, clip]{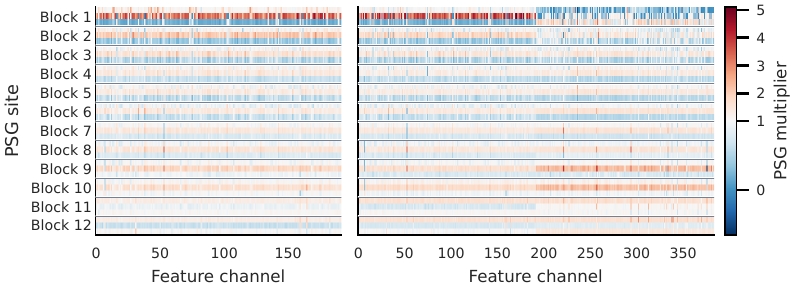}}\hfill
  \caption{(a)~Ablation of different resolution schedules. Two-round variants use $3\!\rightarrow\!6$ heads, while the three-round variant uses $2\!\rightarrow\!4\!\rightarrow\!6$. The $192\!\rightarrow\!240$ schedule offers the best trade-off. (b)~Heatmap of learned PSG multipliers across layers and rounds. Each block shows attention, MLP, and output gates. PSG learns distinct patterns across rounds, layers, and channels.
  }
  \label{fig:psg-profile}
\end{figure*}

\section{\name}

This section describes the \name pipeline. We first present the adaptive-width rounds, followed by cross-resolution token projection, PSG, joint training, and early stopping metrics.

\subsection{Adaptive-Width Progressive Rounds}

\name begins by resizing the input image to a lower resolution and processing it with a narrow subnetwork. Let $\{(r_s,h_s,d_s)\}_{s=1}^{S}$ denote each round, where $r_s$ is the image resolution, $h_s$ is the number of active attention heads, and $d_s$ is the active embedding width. Round 1 uses $(r_1,h_1,d_1)$ and produces a complete prediction. Round 2 process a higher-resolution version of the same image using the wider operating point $(r_2,h_2,d_2)$. Each round traverses all $D$ blocks of the same ViT backbone.

\paragraph{Subnetwork extraction.} We slice the subnetworks following the slicing scheme of adaptive-width models \cite{haberer2024hydravit}. Specifically, to create each subnetwork, we select the first $h_s$ attention heads and their corresponding first $d_s$ channels. We use the same channel prefix throughout the shared Transformer, including the normalization parameters, attention projections, and MLP. At round $s+1$, we resize the input image to $r_{s+1}$, re-embed it, mix the resulting tokens with tokens from previous rounds, and process them using the wider prefix defined by $h_{s+1}$ and $d_{s+1}$; see Figure~\ref{fig:overview}. Because the patch grid changes with the resolution, we interpolate the spatial component of the shared positional embedding to the grid of each round before selecting the active channel prefix.

\paragraph{Cross-Resolution Token Projection.}
Successive rounds reuse tokens from earlier rounds, but differ in token count and embedding width. To reconcile these mismatched representations, we introduce a token projector that resizes the patch-token grid and projects it to the next round's width, while a separate projection adjusts the class token. We then fuse the aligned tokens with fresh, higher-resolution embeddings for the next round.
See \appref{sec:appendix-token-grid} for more information.

\subsection{\textbf{P}rogress-Conditioned \textbf{S}oft \textbf{G}ating (PSG)}

In \name, each Transformer block serves multiple subnetworks across successive inference rounds and therefore processes representations with different resolutions and active widths.
Without stage conditioning, the same parameters must simultaneously optimize representations from different distributions caused by changes in token count, spatial resolution, and active width.
Motivated by iteration-aware modulation \cite{jacobs2026raptor}, we introduce \textbf{P}rogress-Conditioned \textbf{S}oft \textbf{G}ating to condition each block on the progress of inference. Let $X$ denote the input to block $b$ in round $s$. PSG applies channel-wise multipliers to the attention update, MLP update, and final block output; see Figure~\ref{fig:overview}. We summarize the resulting block computation as follows, with blue highlighting the terms introduced by PSG:
\begin{equation}
\begin{aligned}
&X \leftarrow X
+\,\textcolor{progressblue}{
\boldsymbol{G_{s,b}^{\mathrm{attn}}\odot}
\operatorname{\mathbf{LayerScale}}
\!\left(
\textcolor{black}{\operatorname{Att}(X)}
\right)},\\
&X \leftarrow X
+\,\textcolor{progressblue}{
\boldsymbol{G_{s,b}^{\mathrm{mlp}}\odot}
\operatorname{\mathbf{LayerScale}}
\!\left(
\textcolor{black}{\operatorname{MLP}(X)}
\right)},\\
&X_{\mathrm{out}} =
\textcolor{progressblue}{
\boldsymbol{G_{s,b}^{\mathrm{out}}\odot}}\,X.
\end{aligned}
\end{equation}

PSG constructs a five-value metadata vector from the round index, normalized progress across all block executions, the current and previous resolutions, and their relative change. A small shared encoder maps this metadata to a condition embedding. Separate gate heads then produce multiplier vectors for the attention, MLP, and block-output pathways.
Each gate predicts a residual $\Delta G$ and outputs $G=1+\Delta G$. Zero-initializing the final layer makes PSG an identity at initialization, after which it learns channel-wise amplification and suppression across inference stages.

We apply the same conditioning mechanism when combining information across rounds; see Figure~\ref{fig:overview}. For a later round, let $E_s$ denote the new tokens extracted from the higher-resolution image, and let $\widehat{Z}_s$ denote the aligned state from the preceding round. Separate PSG multipliers scale the two streams before fusion:
\begin{equation}
X_s^{(0)}
=
\textcolor{progressblue}{
\boldsymbol{G}_s^{\mathrm{img}} \odot}  E_s
+
\textcolor{progressblue}{
\boldsymbol{G}_s^{\mathrm{prev}} \odot} \widehat{Z}_s
.
\end{equation}
For further details on the PSG architecture, see \appref{sec:appendix-architecture}.

\subsection{Training and Inference}

During training, \name jointly optimizes all rounds using the same classification-loss weight. Early exiting is disabled, so every round is trained on every sample. During inference, after each round, we compute the entropy of the top-10 class probabilities as the uncertainty score. Samples with entropy below a specified threshold exit, while the remaining samples continue to the next round. Varying this threshold produces an accuracy--compute frontier without retraining. At matched continuation rates, the learned router improves accuracy by at most 0.21 percentage points over entropy-based routing; see \Appref{sec:appendix-router}. Given this marginal gain, we favor entropy routing because it is simple and incurs no additional overhead.

\section{Experiments}

\subsection{Setup}

\paragraph{Image classification.}
We evaluate \name on ImageNet-1K \cite{ILSVRC15}. Our implementation uses the DeiT-S architecture \cite{touvron2021training} available in \texttt{timm} \cite{rw2019timm} and contains 24.7M parameters. We train for 300 epochs followed by a 10-epoch cooldown. We report top-1 accuracy on the ImageNet-1K validation set and evaluate the same model without fine-tuning on ImageNet-V2, -A, -R, -Sketch, and -C \cite{recht2019imagenet,hendrycks2019nae,hendrycks2020many,wang2019learning,hendrycks2019benchmarking}. 

\paragraph{DINO representations.}
We apply \name to DINO \cite{caron2021emerging} and train it on ImageNet-1K. Both \name and the fixed ViT-S/16 reference are trained from scratch for 100 epochs using the same DINO setup.

\paragraph{ADE20K segmentation.}
We apply \name to the Segmenter architecture \cite{strudel2021segmenter} and train the resulting model on ADE20K \cite{zhou2017scene} using a linear decoder. 

\begin{figure*}[!t]
  \centering
  \begin{subfigure}[t]{.33\textwidth}
    \centering
    \includegraphics[width=\linewidth]{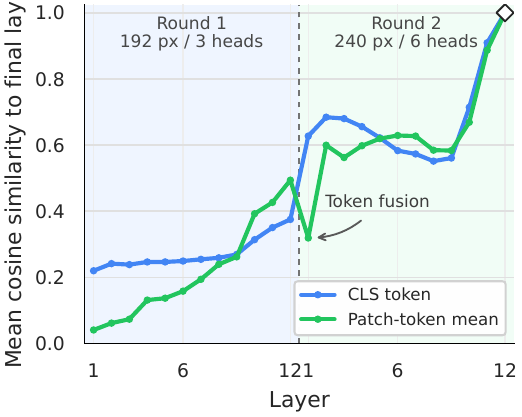}
    \caption{Similarity to round 2's last layer}
    \label{fig:final-similarity}
  \end{subfigure}\hfill
  \begin{subfigure}[t]{.31\textwidth}
    \centering
    \includegraphics[width=\linewidth]{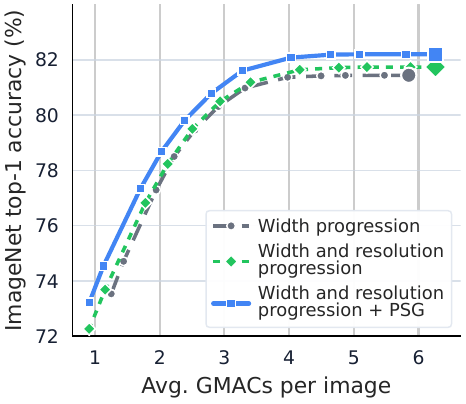}
    \caption{Model construction.}
    \label{fig:construction-components}
  \end{subfigure}\hfill
  \begin{subfigure}[t]{.32\textwidth}
    \centering
    \includegraphics[width=\linewidth]{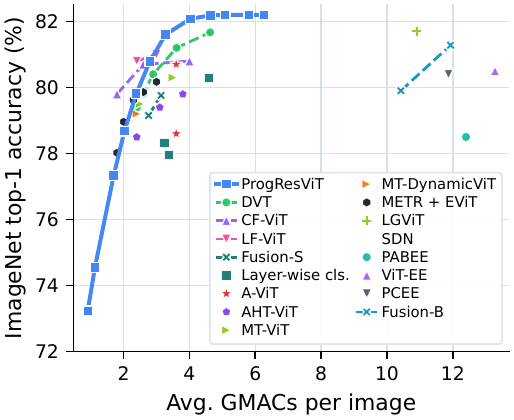}
    \caption{Adaptive-inference baselines.}
    \label{fig:deit_early_exit_comparison}
  \end{subfigure}
  \caption{(a)~Similarity to the final representation generally rises across both rounds, showing that round 2 refines the representation developed in round 1. The patch-token decrease at the round transition reflects fusion with fresh higher-resolution tokens. (b)~Step-by-step ablation of model construction. (c)~Adaptive-inference and early-exit comparison. \name provides the strongest high-accuracy frontier.
}
  \label{fig:reuse-diagnostics}
\end{figure*}

\begin{figure*}[!t]
  \centering
  \subcaptionbox{Round-1 low-pass intervention.\label{fig:construction-frequency}}[.29\textwidth]{%
    \includegraphics[width=\linewidth]{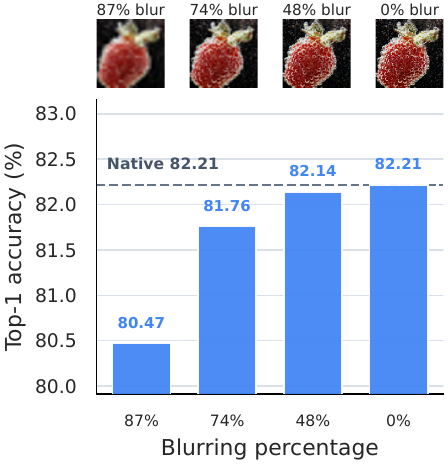}}\hfill
  \subcaptionbox{Token-swap interventions.\label{fig:semantic-token-swap}}[.69\textwidth]{%
    \includegraphics[width=\linewidth]{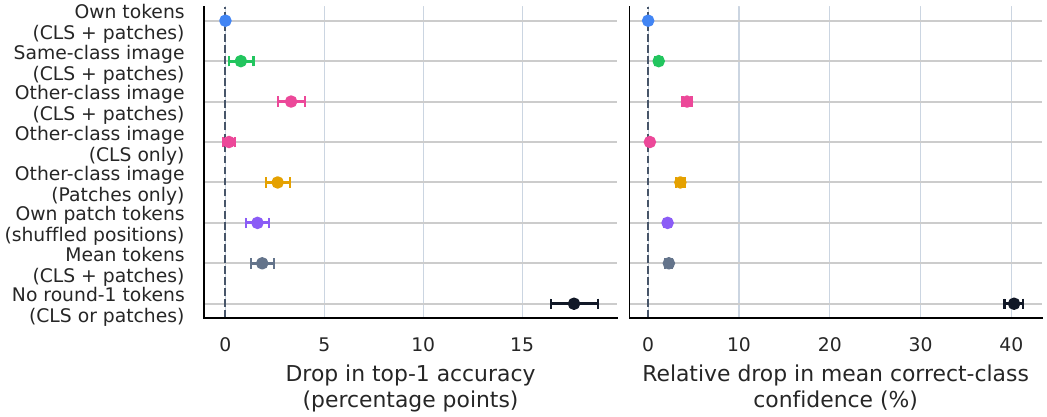}}
  \caption{(a)~Effect of low-pass filtering the round 1 input on final accuracy. Removing high-frequency information from round 1 reduces round 2 accuracy, indicating that the first-round representation provides useful context for refinement.
(b)~Effect of replacing or shuffling round 1 tokens on round 2 accuracy and correct-class confidence. Perturbing the patch tokens causes the largest drop, indicating that they carry useful semantic and spatial information.}
  \label{fig:construction}
\end{figure*}

\begin{table}[t]
\centering
\small
\begin{tabular}{lccc}
\toprule
Schedule & Round 1 & Round 2 & Routed\\
\midrule
$192\!\rightarrow\!240$ & 73.23/.91 & 82.21/6.27 & 82.18/4.47\\
$160\!\rightarrow\!240$ & 70.95/.62 & 82.14/5.97 & 82.14/4.96\\
$160\!\rightarrow\!384$ & 70.62/.62 & 83.70/16.15 & 83.69/13.29\\
\midrule
$192\!\rightarrow\!240$ + KD & 76.02/.91 & 83.80/6.27 & 83.77/4.46\\
$160\!\rightarrow\!384$ + KD & 73.92/.62 & 84.90/16.15 & 84.87/11.12\\
\bottomrule
\end{tabular}

\caption{Resolution schedules and knowledge-distillation (KD) evaluation. Each cell reports top-1 accuracy / average GMACs. Routed accuracy denotes an efficient operating point with accuracy close to round 2.}
\label{tab:schedules}
\end{table}

\begin{figure*}[!t]
  \centering
  \subcaptionbox{Dynamic token baselines.\label{fig:deployment-dynamic}}[.325\textwidth]{%
    \includegraphics[width=\linewidth]{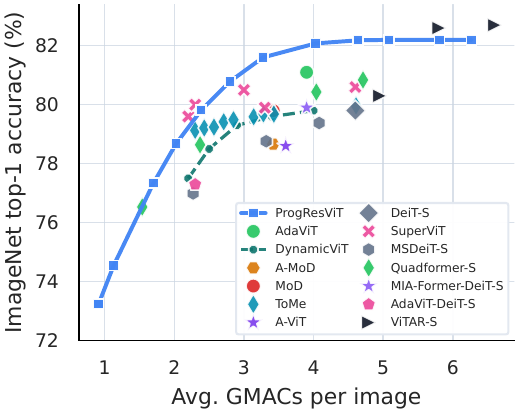}}\hfill
  \subcaptionbox{Throughput frontier.\label{fig:deployment-throughput}}[.325\textwidth]{%
    \includegraphics[width=\linewidth]{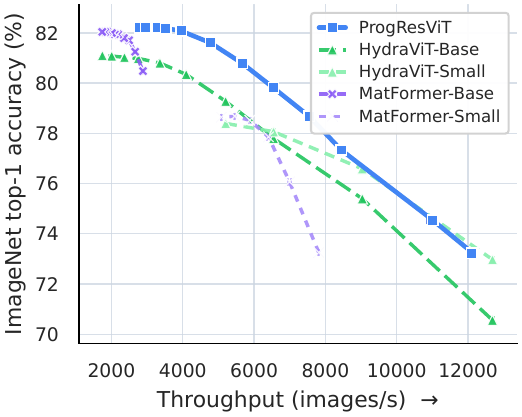}}\hfill
  \subcaptionbox{Adaptive-width baselines.\label{fig:full-accuracy-compute-panel}}[.325\textwidth]{%
    \includegraphics[width=\linewidth]{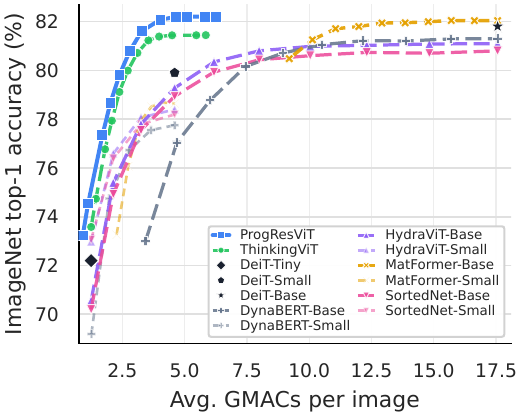}}
\caption{(a)~Dynamic-token comparison. \name provides the strongest high-accuracy frontier; lines denote one trained model, while isolated markers denote separate models. (b)~Throughput comparison on one NVIDIA L40 at batch size 512. \name achieves higher accuracy at comparable throughput. (c)~Adaptive-width comparison. \name outperforms the evaluated baselines at comparable compute.}
  \label{fig:deployment}
\end{figure*}

\subsection{Accuracy--Efficiency Trade-offs for classification}

\paragraph{Resolution and width scheduling.} 
Figure~\ref{fig:schedules} compares the accuracy--compute frontiers obtained with different resolution and width schedules. Among the tested schedules, the configuration using a $192\!\times\!192$ input in the first round and a $240\times240$ input in the second round, denoted by $192\!\rightarrow\!240$, together with an attention-head schedule of 3 heads in the first round and 6 heads in the second round, denoted by $3\!\rightarrow\!6$, provides the strongest overall trade-off. We therefore use it as the default configuration for the remaining analyses and ablations. The $160\rightarrow384$ with a $3\!\rightarrow\!6$ head schedule requires more computation but gives the highest final accuracy.

Table~\ref{tab:schedules} reports the round 1, round 2, and routed results with head schedule of $3\!\rightarrow\!6$. Routed accuracy denotes an efficient operating point with accuracy close to round 2. For the default $192\!\rightarrow\!240$ configuration, round 1 achieves 73.23\% accuracy at 0.91 GMACs, while the round 2 reaches 82.21\% accuracy at 6.27 GMACs. Entropy-based routing retains 82.18\% accuracy at 4.47 average GMACs, reducing computation by 28.7\% with a 0.03-point accuracy drop. The table also includes the higher-accuracy $160\!\rightarrow\!384$ configuration, which achieves 83.70\% after the second round. \appref{sec:appendix-fixed-resolution} presents results for DeiT-S at different resolutions.

To maximize performance, we also train \name for 300 epochs using knowledge distillation from a DeiT-III-B/384 teacher~\cite{touvron2022deit}. The $160\!\rightarrow\!384$ model achieves 84.90\% at 16.15 GMACs and retains 84.87\% at 11.12 GMACs.
Despite using a DeiT-S architecture, the $160\!\rightarrow\!384$ configuration achieves 84.90\% top-1 accuracy, slightly above the reported 84.80\% accuracy of ImageNet-21K-pretrained DeiT-III-S/384 at 15.5 GMACs.
For the $192\!\rightarrow\!240$ configuration, distillation increases the round 1 accuracy from 73.23\% to 76.02\% and the round 2 accuracy from 82.21\% to 83.80\%. With entropy-based routing, the distilled model retains 83.77\% at 4.46 GMACs.
Distillation details are provided in \appref{sec:appendix-distillation}.

\begin{figure*}[!t]
  \centering

  \subcaptionbox{PSG multiplier distributions for the attention, MLP, and layer-output.\label{fig:gates-pca-gates}}[.62\textwidth]{%
    \includegraphics[width=\linewidth]{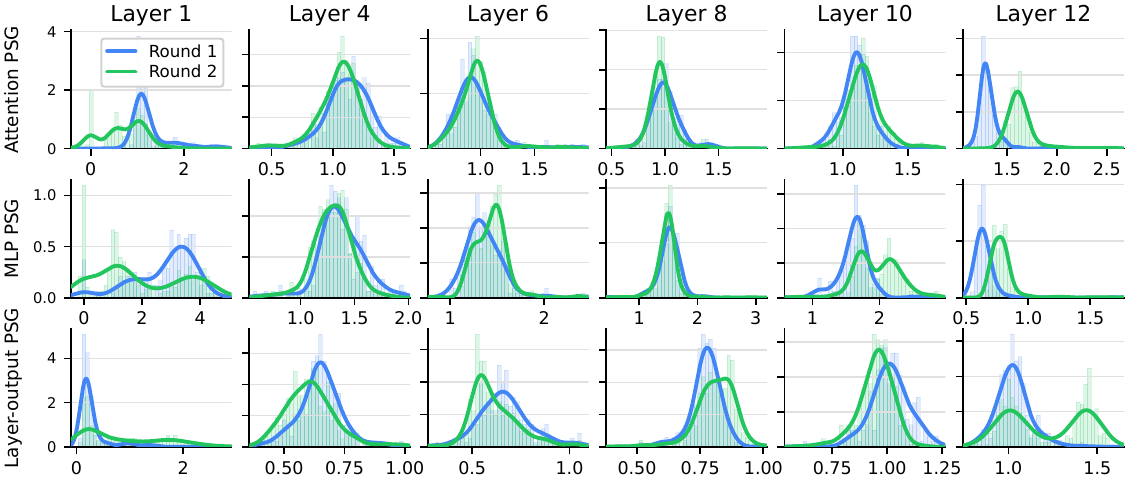}}
  \hfill
  \subcaptionbox{PCA of the CLS trajectories.\label{fig:gates-pca-pca}}[.33\textwidth]{%
    \includegraphics[width=\linewidth]{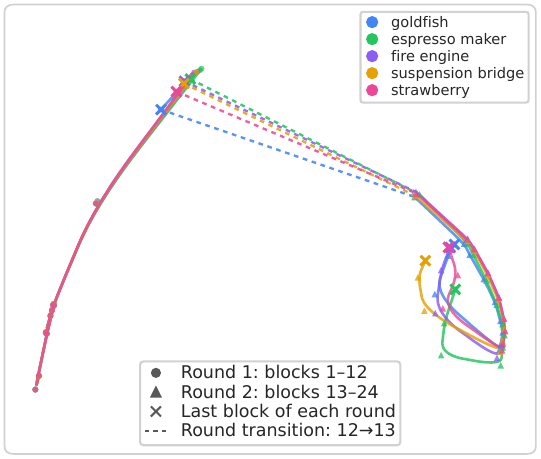}}

  \medskip

  \subcaptionbox{Performance of \name and DeiT models across five hard ImageNet variants.%
    \label{fig:robustness-variants}}[\textwidth]{%
    \includegraphics[width=.97\linewidth]{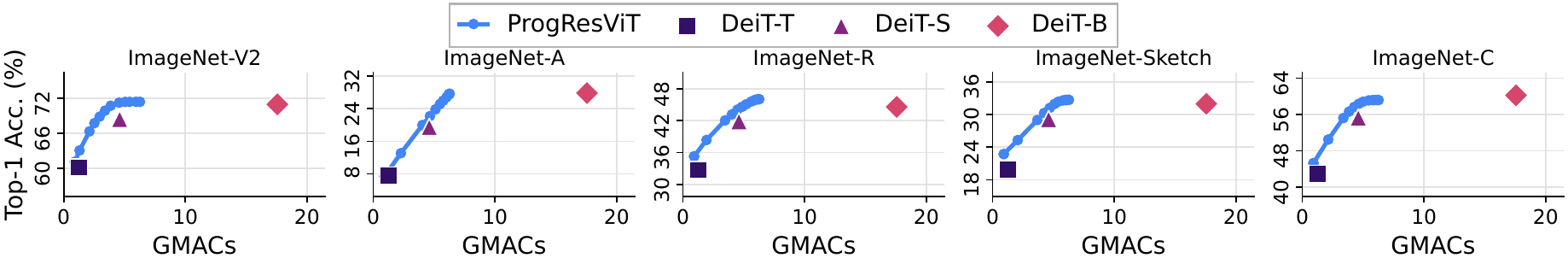}}

  \caption{%
    (a)~PSG multipliers at selected layers, with the strongest
    round-specific separation in the final block.
    (b)~PCA trajectories of five class-mean CLS representations.
    Separation begins in round 1 and strengthens in round 2; the dashed
    line marks token fusion.
    (c)~Robustness on five hard ImageNet variants. \name maintains
    favorable trade-offs over DeiT baselines.}
  \label{fig:gates-pca}
\end{figure*}

To evaluate \name under challenging distribution shifts, we test it without fine-tuning on ImageNet-V2, -A, -R, -Sketch, and -C, and report in Figure~\ref{fig:robustness-variants}. These benchmarks cover natural distribution shift, adversarially filtered images, artistic renditions, sketches, and common corruptions, making them substantially more challenging than standard ImageNet-1K. \name maintains efficient accuracy--compute trade-offs across all five datasets, outperforming DeiT-S throughout and DeiT-B on several variants despite using substantially less compute, which shows the effectiveness of \name's routing mechanism.

\paragraph{Adaptive-width baselines.}
Figure~\ref{fig:full-accuracy-compute-panel} compares \name with adaptive-width baselines across the accuracy–GMACs trade-off points~\cite{haberer2024hydravit,devvrit2024matformer,hou2020dynabert,valipour2023sortednet}. \name outperforms these adaptive-width baselines. Furthermore, \name outperforms ThinkingViT, its primary baseline, across all operating points, achieving gains of up to 0.74 percentage points~\cite{hojjat2026thinkingvit}. These results demonstrate the effectiveness of the proposed PSG and \name's progressive resolution-width inference.
Figure~\ref{fig:deployment-throughput} shows a similar trend for measured throughput. Comparisons with FlexiViT~\cite{beyer2023flexivit} and memory measurements are provided in \appref{sec:appendix-flexivit} and \appref{sec:appendix-memory}, respectively.

\paragraph{Dynamic-token and early-exit baselines.}
As \name adapts resolution and width while preserving the underlying ViT block structure and retaining the full token grid and depth within each round, it is complementary to dynamic-token and depth-wise early-exit methods. These mechanisms could therefore be incorporated within individual rounds. Nevertheless, we compare against both categories: Figures~\ref{fig:deployment-dynamic} and~\ref{fig:deit_early_exit_comparison} show that \name provides the strongest high-accuracy frontier among the baselines. Further details and references are provided in \appref{sec:appendix-adaptive-baselines}.

\subsection{Transfer Beyond Supervised Classification}

\paragraph{Self-supervised representations (DINO).}
We apply \name to a DINO ViT-S/16 backbone and train it alongside fixed-resolution $224\times224$ and $240\times240$  baselines for 100 epochs using matched recipes \cite{caron2021emerging}. Under 20-NN evaluation, \name outperforms both separately trained ViT-S/16 baselines; see Figure~\ref{fig:main-frontier-dino}. These results show that \name extends effectively to self-supervised representations while preserving accuracy--compute trade-offs. Linear-probing results and attention maps are provided in \appref{sec:appendix-dino} and \appref{sec:appendix-attention-maps}, respectively.

\paragraph{Semantic segmentation.}
We replace Segmenter's fixed ViT encoder with \name and use one shared linear decoder for both rounds. Round 1 and 2 processes a $448\times448$ and $576\times576$ input respectively. \name outperforms ThinkingViT and the separately trained DeiT-Tiny, DeiT-Small, and DeiT-Base segmentation models; see Figure~\ref{fig:main-frontier-seg}. Segmentation-routing details are provided in \appref{sec:appendix-segmentation}.

\subsection{Analysis of \textbf{P}rogress-Conditioned \textbf{S}oft \textbf{G}ating}

To investigate how PSG adapts the shared backbone, Figure~\ref{fig:psg-profile-values} visualizes its channel-wise attention, MLP, and block-output multipliers across all blocks and rounds. Their variation across channels, depth, and rounds under a shared color scale shows that PSG learns stage-specific amplification and suppression rather than uniform scaling. To further investigate this behavior, we compare its multiplier distributions at selected depths in Figure~\ref{fig:gates-pca-gates}. Round-specific differences appear throughout the backbone, showing that PSG learns nontrivial, stage-dependent modulation.

\subsection{Progressive Refinement}

\paragraph{Ablation of \name's construction} Figure~\ref{fig:construction-components} compares the main stages of the model design. The fixed-resolution width baseline with $3\!\rightarrow\!6$ head schedule achieves 81.44\% accuracy. Introducing the $192\!\rightarrow\!240$ resolution progression increases the full-path accuracy to 81.74\%. Finally, adding PSG improves the accuracy to 82.20\% at nearly the same GMACs.

\paragraph{Round 1 Provides Useful Context for Round 2.}
To assess how information from the low-resolution first round contributes to the final prediction, we apply a low-pass intervention to the round 1 input and measure the resulting round 2 accuracy; see Figure~\ref{fig:construction-frequency}. Specifically, we remove high-frequency information by blurring the round 1 image while leaving the round 2 input unchanged. The horizontal axis reports the blurring percentage, with 0\% denoting the native image. At approximately 87\% blurring, the accuracy of the second round drops by 1.74\%. This reduction indicates that the representations produced in round 1 provide meaningful information that supports prediction in round 2.
Figure~\ref{fig:semantic-token-swap} examines the contribution of the round 1 representation from another perspective. We replace the round 1 tokens with tokens from other images while keeping the round 2 image embeddings fixed. Replacing or shuffling the patch tokens reduces both round 2 accuracy and correct-class confidence, showing that they carry useful semantic and spatial information. Replacing only the class token has the smallest effect, especially when it comes from an image of the same class, as class tokens from similar images are expected to contain similar information.
For more analysis, see \appref{sec:appendix-replacement-accuracy} and \appref{sec:appendix-replacement-confidence}, respectively.

\paragraph{Progressive Representation Refinement Across Rounds.}
Figure~\ref{fig:final-similarity} shows that block outputs become increasingly similar to the final round 2 representation, indicating that round 2 refines rather than restarts the representation learned in round 1. Patch-token similarity briefly decreases at the transition as fusion introduces fresh higher-resolution tokens, whereas CLS-token similarity continues to increase across both rounds. Figure~\ref{fig:gates-pca-pca} shows PCA trajectories of class-mean CLS representations for five ImageNet classes. Round 1 moves them from a shared region toward partial separation, while the dashed transition marks token fusion. Round 2 further separates the trajectories, indicating that the higher-resolution stage strengthens class-specific representations.

\section{Limitations}

Although \name reduces computation for easy images, it may spend additional computation on uncertain inputs that remain misclassified after the final round. This can occur when uncertainty reflects ambiguity or distribution shift rather than insufficient computation. The current entropy-based router does not explicitly distinguish such cases from inputs that benefit from refinement. Although rejection mechanisms can reduce this unnecessary computation, incorporating them compromises the fairness of the comparison with other baselines.

\section{Conclusion}

We introduced \name, an input-adaptive ViT that progressively increases input resolution and active model width while reusing features across rounds. A cross-resolution token projector aligns features between stages, and PSG conditions token fusion and shared Transformer blocks on the current inference stage. On ImageNet-1K, \name improves accuracy--compute trade-offs over adaptive-width, adaptive-depth, and dynamic-token baselines, while the distilled model reaches 84.9\% top-1 accuracy. The same design also transfers effectively to DINO representation learning and ADE20K semantic segmentation.

\section*{Acknowledgments}

This research received funding from the Federal Ministry for Digital and Transport under the CAPTN-F\"orde 5G project (grant no.~45FGU139H), the German Ministry of Transport and Digital Infrastructure through the CAPTN F\"orde Areal II project (grant no.~45DTWV08D), the Federal Ministry for Economic Affairs and Energy under the CAPTN X-FERRY project (grant no.~03SX612A), and the Federal Ministry for Economic Affairs and Climate Action under the Marispace-X project (grant no.~68GX21002E). The work was supported in part by high-performance computing resources provided by the Kiel University Computing Centre and the Hydra computing cluster, funded by the German Research Foundation (grant no.~442268015) and the Petersen Foundation (grant no.~602157).

\bibliography{aaai2027}

\clearpage
\appendix
\setcounter{secnumdepth}{1}
\setcounter{section}{0}
\setcounter{figure}{0}
\setcounter{table}{0}
\setcounter{equation}{0}
\renewcommand{\thesection}{\Alph{section}}
\renewcommand{\thefigure}{S\arabic{figure}}
\renewcommand{\thetable}{A\arabic{table}}
\renewcommand{\theequation}{S\arabic{equation}}
\section*{Appendix}
\suppressfloats[t]

\lstdefinestyle{psgpseudocode}{
  language=Python,
  basicstyle=\ttfamily\scriptsize,
  keywordstyle=\bfseries,
  commentstyle=\itshape,
  columns=fullflexible,
  keepspaces=true,
  breaklines=true,
  breakatwhitespace=true,
  showstringspaces=false,
  xleftmargin=2pt,
  xrightmargin=2pt,
  aboveskip=5pt,
  belowskip=5pt
}

\section{Token Projector Pipeline}
\label{sec:appendix-token-grid}

The token projector aligns the previous-round representation with the next round in both grid size and channel width. It processes the class token separately from the patch tokens. The patch tokens are reshaped to their spatial grid, bilinearly resized, refined by an identity-initialized depthwise $3\!\times\!3$ convolution, and expanded by a learned $1\!\times\!1$ projection. A separate linear layer expands the class token before all tokens are concatenated again. For the default $192^2\!\rightarrow\!240^2$ model, this maps a $12\!\times\!12$ grid with 192 channels to a $15\!\times\!15$ grid with 384 channels.

\begin{lstlisting}[style=psgpseudocode]
class TokenProjector(nn.Module):
    def __init__(self, d_in, d_out, source_hw, target_hw):
        self.source_hw, self.target_hw = source_hw, target_hw
        self.depthwise = Conv2d(
            d_in, d_in, kernel_size=3, padding=1,
            groups=d_in, bias=False)
        self.channel_proj = Conv2d(d_in, d_out, kernel_size=1)
        self.class_proj = Linear(d_in, d_out)
        initialize_as_identity(self.depthwise)
        initialize_shared_channels_as_identity(
            self.channel_proj, self.class_proj)

    def forward(self, tokens):
        class_token = tokens[:, :1]
        patch_tokens = tokens[:, 1:]
        patch_grid = reshape_to_grid(patch_tokens, self.source_hw)
        patch_grid = interpolate(
            patch_grid, size=self.target_hw,
            mode="bilinear", align_corners=False)
        patch_grid = self.depthwise(patch_grid)
        patch_grid = self.channel_proj(patch_grid)
        patch_tokens = flatten_to_tokens(patch_grid)
        class_token = self.class_proj(class_token)
        return concatenate([class_token, patch_tokens], dim=1)
\end{lstlisting}

\section{PSG Architecture}
\label{sec:appendix-architecture}
\suppressfloats[t]

PSG uses five metadata values: the round index, normalized progress through all round--block applications, current resolution, previous resolution, and resolution ratio. A shared metadata encoder maps these values to a conditioning vector. Five branch-specific heads then produce channel-wise residual multipliers for the new-image stream, projected previous-round stream, attention update, MLP update, and complete block output. The first two multipliers control token fusion at a round transition, while the remaining three modulate every shared Transformer block. The PyTorch-style pseudocode below summarizes this architecture.

\begin{lstlisting}[style=psgpseudocode]
class PSG(nn.Module):
    branches = ["image", "previous", "attention",
                "mlp", "block"]
    def __init__(self, max_width, rounds, blocks):
        self.S, self.D = rounds, blocks
        self.encoder = Sequential(Linear(5, 128), SiLU())
        self.heads = ModuleDict({
            b: Sequential(Linear(128, 128), SiLU(),
                          Linear(128, max_width))
            for b in self.branches})
        zero_init(last_linear(self.heads))
        self.attn_scale = Parameter(ones(max_width))
        self.mlp_scale = Parameter(ones(max_width))
    def metadata(self, s, b, resolution, previous_res):
        previous_res = resolution if s == 0 else previous_res
        progress = (s * self.D + b) / (self.S * self.D - 1)
        return tensor([s, progress, log2(resolution / 224),
                       log2(previous_res / 224),
                       log2(resolution / previous_res)])
    def multipliers(self, metadata, width):
        condition = self.encoder(metadata)
        return {b: 1 + head(condition)[:width]
                for b, head in self.heads.items()}
    def fuse(self, new_tokens, previous_tokens, gate):
        return (gate["image"] * new_tokens
                + gate["previous"] * previous_tokens)
    def transformer_block(self, x, gate):
        d = x.shape[-1]
        x += gate["attention"] * self.attn_scale[:d] \
             * attention_update(x)
        x += gate["mlp"] * self.mlp_scale[:d] * mlp_update(x)
        return gate["block"] * x
\end{lstlisting}

The last linear layer of every conditioned head is zero-initialized, so all PSG multipliers initially equal one. The PSG-specific attention and MLP scales are also identity-initialized and are distinct from backbone LayerScale.

\section{Entropy Routing Versus a Learned Router}
\label{sec:appendix-router}

We test whether a learned continuation rule can select more useful images for refinement than prediction entropy. This experiment uses the $192\!\rightarrow\!240$ checkpoint without knowledge distillation. The auxiliary router receives 23 summary statistics computed from the frozen round-1 logits, including top-10 and full entropy, confidence and logit margins, distribution moments, and the leading probabilities and logits. A $23\!\rightarrow\!256\!\rightarrow\!128\!\rightarrow\!1$ MLP with 39,681 trainable parameters is fitted for six epochs on 50,000 ImageNet training images to predict whether round 2 corrects a round-1 error, then evaluated on all 50,000 validation images.

Table~\ref{tab:learned-router} reports GMACs and accuracy separately for entropy and the learned router using their corresponding evaluation records. In every row, both methods send the same fraction of images to round 2.
The learned router improves selection most at constrained budgets. Its largest measured gain is 0.214 percentage points when 20\% of images continue, and the gain remains approximately 0.2 points through the 20--40\% region.
It falls to 0.036 points at 50\% continuation, 0.012 points at 60\%, and shows no consistent advantage thereafter. Learning can therefore improve routing modestly, but the benefit is localized and small. Entropy requires no auxiliary training set, learned parameters, feature normalization, or additional deployment path, and therefore we use entropy as the simpler default.
\begin{table}[t]
\centering
\setlength{\tabcolsep}{2.15pt}
\begin{tabular}{rccccc}
\toprule
& \multicolumn{2}{c}{Entropy} & \multicolumn{2}{c}{Learned router} & \\
\cmidrule(lr){2-3}\cmidrule(lr){4-5}
Continue & GMACs & Top-1 & GMACs & Top-1 & $\Delta$\\
(\%) & / image & (\%) & / image & (\%) & (p.p.)\\
\midrule
0   & 0.912 & 73.234 & 0.912 & 73.234 & $+0.000$\\
10  & 1.447 & 76.234 & 1.447 & 76.334 & $+0.100$\\
20  & 1.983 & 78.518 & 1.983 & 78.732 & $+0.214$\\
30  & 2.518 & 80.218 & 2.518 & 80.394 & $+0.176$\\
40  & 3.054 & 81.234 & 3.054 & 81.438 & $+0.204$\\
50  & 3.590 & 81.900 & 3.590 & 81.936 & $+0.036$\\
60  & 4.125 & 82.120 & 4.125 & 82.132 & $+0.012$\\
70  & 4.661 & 82.200 & 4.661 & 82.192 & $-0.008$\\
80  & 5.196 & 82.212 & 5.196 & 82.212 & $+0.000$\\
90  & 5.732 & 82.210 & 5.732 & 82.210 & $+0.000$\\
100 & 6.267 & 82.210 & 6.267 & 82.210 & $+0.000$\\
\bottomrule
\end{tabular}
\caption{Entropy and learned routing at matched continuation rates on ImageNet-1K. GMACs and top-1 accuracy are listed separately from the corresponding routing evaluations; $\Delta$ is learned minus entropy accuracy. The equal backbone GMACs follow from matching the number of images refined. Learned-router overhead is excluded.}
\label{tab:learned-router}
\end{table}

\section{Fixed-Resolution Deployment Context}
\label{sec:appendix-fixed-resolution}

\begin{figure}[t]
\centering
\includegraphics[width=.82\columnwidth]{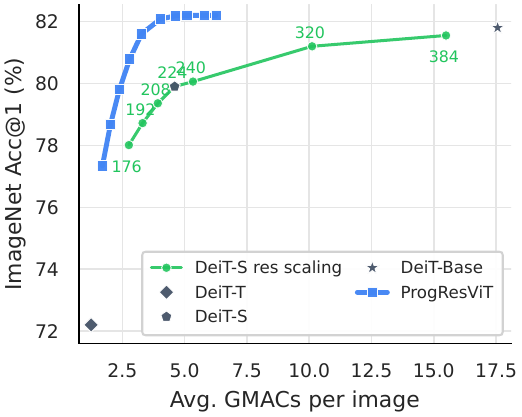}
\caption{Joint resolution and width progression compared with available fixed-resolution DeiT records.}
\label{fig:fixed-resolution-context}
\end{figure}

Figure~\ref{fig:fixed-resolution-context} places the progressive operating points beside fixed-model resolution scaling. The available DeiT-S record at $384^2$ reaches 81.55\% at 15.49 GMACs, whereas the full \name path reaches 82.21\% at 6.27 GMACs. This is 0.66 percentage points higher with 59.5\% less analytical compute.

\section{Knowledge-Distillation Details}
\label{sec:appendix-distillation}

We train the distilled variants for 300 epochs using hard logit-level distillation from a frozen DeiT-III-B/384 teacher pretrained on ImageNet-21K and fine-tuned on ImageNet-1K. For each augmented image, the teacher processes the $384^2$ view and provides the hard pseudo-label $\hat{y}_t=\arg\max z_t$. The loss for round $s$ is
\[
\mathcal{L}_s
=0.5\,\mathrm{CE}(z_s,y)
+0.5\,\mathrm{CE}(z_s,\hat{y}_t),
\qquad
\mathcal{L}=\tfrac{1}{2}\bigl(\mathcal{L}_1+\mathcal{L}_2\bigr).
\]
The teacher remains in evaluation mode and receives no gradients. Because the target is a hard class label, no temperature parameter is used. 

\section{Near-Lossless Routing Points}
\label{sec:appendix-routing-points}
\label{sec:appendix-frontiers}

We select the lowest-compute entropy threshold whose top-1 accuracy remains within 0.03 percentage points of full-path inference. Table~\ref{tab:best-tradeoff} summarizes the resulting operating points. They retain essentially the full accuracy while reducing average computation by 21.9--31.1\%. Table~\ref{tab:schedules-supp} provides the complete threshold sweep for every finished two-round classification model. Each entry is computed from the full 50,000-image ImageNet-1K validation set.

\begin{table}[t]
\centering
\setlength{\tabcolsep}{3.2pt}
\begin{tabular}{lrrrrr}
\toprule
Model & Full & $\tau$ & Routed & Routed & Saving\\
& top-1 & & top-1 & GMACs & (\%)\\
\midrule
$192\!\rightarrow\!240$ & 82.206 & 0.3505 & 82.178 & 4.467 & 28.7\\
$192\!\rightarrow\!240$ + KD & 83.796 & 0.20923 & 83.766 & 4.461 & 28.8\\
$160\!\rightarrow\!384$ & 83.700 & 0.2508 & 83.672 & 12.606 & 21.9\\
$160\!\rightarrow\!384$ + KD & 84.902 & 0.22911 & 84.872 & 11.124 & 31.1\\
\bottomrule
\end{tabular}
\caption{Near-lossless ImageNet operating points. Each routed point minimizes compute while remaining within 0.03 percentage points of its full-path top-1 accuracy.}
\label{tab:best-tradeoff}
\end{table}

\begin{table*}[t]
\centering
\setlength{\tabcolsep}{.15pt}
\renewcommand{\arraystretch}{0.96}
\begin{tabular*}{\textwidth}{@{\extracolsep{\fill}}r *{10}{r}@{}}
\toprule
& \multicolumn{2}{c}{$192\!\rightarrow\!240$}
& \multicolumn{2}{c}{$192\!\rightarrow\!240$ + KD}
& \multicolumn{2}{c}{$160\!\rightarrow\!240$}
& \multicolumn{2}{c}{$160\!\rightarrow\!384$}
& \multicolumn{2}{c}{$160\!\rightarrow\!384$ + KD}\\
\cmidrule(lr){2-3}\cmidrule(lr){4-5}\cmidrule(lr){6-7}\cmidrule(lr){8-9}\cmidrule(lr){10-11}
$\tau$ & GMACs & Top-1 & GMACs & Top-1 & GMACs & Top-1 & GMACs & Top-1 & GMACs & Top-1\\
\midrule
0.0  & 6.267 & 82.206 & 6.267 & 83.796 & 5.971 & 82.144 & 16.152 & 83.700 & 16.152 & 84.902\\
0.1  & 5.805 & 82.206 & 5.270 & 83.798 & 5.605 & 82.144 & 15.089 & 83.700 & 13.606 & 84.900\\
0.2  & 5.087 & 82.206 & 4.509 & 83.772 & 4.963 & 82.144 & 13.290 & 83.686 & 11.547 & 84.872\\
0.3  & 4.642 & 82.192 & 4.056 & 83.714 & 4.537 & 82.122 & 12.058 & 83.646 & 10.279 & 84.822\\
0.5  & 4.030 & 82.086 & 3.461 & 83.576 & 3.937 & 81.976 & 10.364 & 83.504 & 8.571 & 84.606\\
0.8  & 3.275 & 81.606 & 2.727 & 82.844 & 3.201 & 81.400 & 8.216 & 82.834 & 6.479 & 83.728\\
1.0  & 2.798 & 80.790 & 2.316 & 82.018 & 2.727 & 80.512 & 6.799 & 81.950 & 5.224 & 82.646\\
1.2  & 2.386 & 79.816 & 1.982 & 81.030 & 2.282 & 79.318 & 5.537 & 80.712 & 4.175 & 81.372\\
1.4  & 2.026 & 78.682 & 1.689 & 80.008 & 1.896 & 77.956 & 4.405 & 79.052 & 3.267 & 79.950\\
1.6  & 1.700 & 77.340 & 1.443 & 78.876 & 1.543 & 76.462 & 3.363 & 77.268 & 2.469 & 78.586\\
2.0  & 1.129 & 74.542 & 1.042 & 76.816 & 0.895 & 72.856 & 1.438 & 72.866 & 1.100 & 75.328\\
5.0  & 0.912 & 73.226 & 0.912 & 76.018 & 0.615 & 70.950 & 0.615 & 70.624 & 0.615 & 73.920\\
10.0 & 0.912 & 73.226 & 0.912 & 76.018 & 0.615 & 70.950 & 0.615 & 70.624 & 0.615 & 73.920\\
\bottomrule
\end{tabular*}
\caption{Full entropy-routing sweeps for the two-round ImageNet-1K models. GMACs denote average computation per image, and Top-1 is reported in percent.}
\label{tab:schedules-supp}
\end{table*}

\section{Comparison with Resolution-Flexible Inference}
\label{sec:appendix-flexivit}

\begin{figure}[t]
  \centering
  \includegraphics[width=.8\columnwidth]{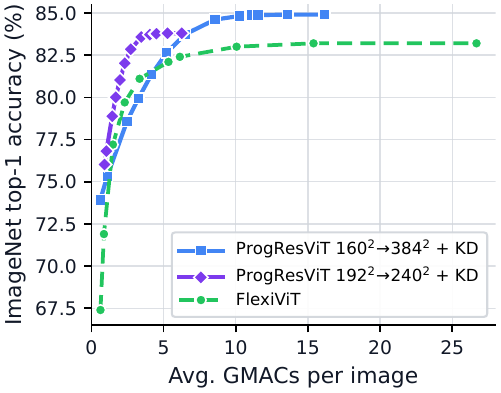}
  \caption{Accuracy--compute comparison with FlexiViT. The \name curves vary the entropy thresholds of the $192^2\!\rightarrow\!240^2$ and $160^2\!\rightarrow\!384^2$ KD models; FlexiViT uses the supplied resolution-flexible operating points.}
\label{fig:flexivit-comparison}
\end{figure}

Figure~\ref{fig:flexivit-comparison} compares the entropy-routed frontiers of the $192^2\!\rightarrow\!240^2$ and $160^2\!\rightarrow\!384^2$ KD models with the FlexiViT accuracy--compute operating points~\cite{beyer2023flexivit}. The methods vary computation differently: \name routes images across progressive rounds, whereas FlexiViT supports inference at multiple patch sizes. The $192^2\!\rightarrow\!240^2$ model reaches 83.766\% at 4.462 GMACs, while the $160^2\!\rightarrow\!384^2$ model reaches 84.902\% at 16.152 GMACs; FlexiViT reaches 83.2\% at 15.39 GMACs.

\section{Deployment Memory}
\label{sec:appendix-memory}

\begin{figure}[t]
\centering
\includegraphics[width=.82\columnwidth]{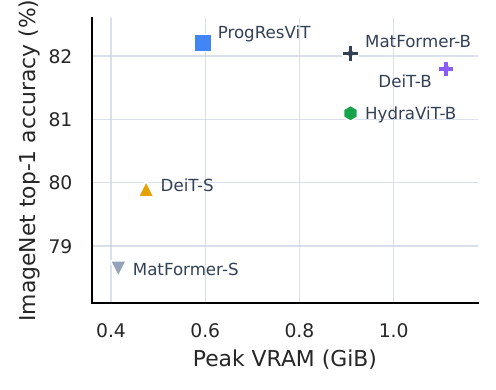}
\caption{Figure~\ref{fig:deployment-memory} reports peak allocated GPU memory at batch size 128 on a single NVIDIA L40 under the same execution setup used for the throughput measurements. For \name, we measure the complete round-1-to-round-2 inference path.}
\label{fig:deployment-memory}
\end{figure}

Figure~\ref{fig:deployment-memory} reports peak allocated GPU memory at batch size 128 on a single NVIDIA L40 under the same execution setup used for the throughput measurements.

\section{Dynamic-Token and Early-Exit Baselines}
\label{sec:appendix-adaptive-baselines}

\ifdefined\PROGRESVITINLINEAPPENDIX Figure~\ref{fig:deployment-dynamic}\else The dynamic-token comparison in the main paper\fi\ compares \name with AdaViT, DynamicViT, A-MoD, MoD, ToMe, A-ViT, SuperViT, MSDeiT-S, QuadFormer-S, MIA-Former, and ViTAR-S, with DeiT-S as a fixed reference \cite{meng2022adavit,rao2021dynamicvit,gadhikar2025attention,raposo2024mixture,bolya2023token,yin2022vit,lin2022supervit,havtorn2023msvit,ronen2023mixed,yu2021miaformer,fan2024vitar,touvron2021training}. \ifdefined\PROGRESVITINLINEAPPENDIX Figure~\ref{fig:deit_early_exit_comparison}\else The early-exit comparison in the main paper\fi\ compares \name with DVT, CF-ViT, LF-ViT, Fusion, a layer-wise classifier, A-ViT, AHT-ViT, Multi-Tailed ViT, METR with EViT, LGViT, SDN, PABEE, ViT-EE, and PCEE \cite{wang2021not,chen2023cfvit,hu2024lfvit,pradeep2026fusion,jiang2025tracing,yin2022vit,shutov2025ahtvit,wang2024multitailed,liu2024metr,liang2022evit,xu2023lgvit,kaya2019shallow,zhou2020bert,bakhtiarnia2021multi,zhang2022pcee}. In both figures, connected points denote operating points from one trained model, whereas isolated markers denote separately trained configurations.

\section{DINO Linear Probing}
\label{sec:appendix-dino}

\begin{figure}[t]
\centering
\includegraphics[width=.8\columnwidth]{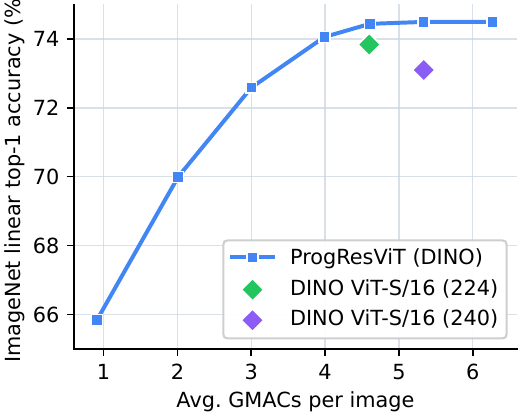}
\caption{DINO linear-probing accuracy--compute comparison. \name outperforms fixed-resolution ViT-S/16 baselines at comparable compute.}
\label{fig:dino-linear}
\end{figure}

Figure~\ref{fig:dino-linear} reports linear-probing accuracy using frozen DINO encoders. At comparable compute, \name outperforms the separately trained fixed-resolution ViT-S/16 baselines at $224^2$ and $240^2$, showing that progressive resolution--width inference also benefits self-supervised representations. \name and DINO baselines are trained for 100 epochs.

\section{ADE20K Segmentation Routing}
\label{sec:appendix-segmentation}

We adapt entropy-based routing to semantic segmentation, where uncertainty must be aggregated across spatial predictions. After Round 1, the shared segmentation decoder produces a class-probability distribution for every patch. We compute the entropy of each distribution, select the most uncertain 10\% of patches, and average their entropies to obtain one uncertainty score for the image. Images whose score exceeds the routing threshold continue to Round 2 for higher-resolution, wider processing; the remaining images exit after Round 1. Varying the threshold controls how many images receive the second-round computation.

\begin{table}[t]
\centering
\small
\setlength{\tabcolsep}{4pt}
\begin{tabular}{@{}p{.70\columnwidth}r@{}}
\toprule
Component & GMACs\\
\midrule
\multicolumn{2}{@{}l}{\textit{Round 1: $192^2$, 3 heads, 192 channels}}\\
ViT path (without PSG) & 0.91\\
Block PSG & 0.002\\
\cmidrule(lr){1-2}
\textbf{Round-1 total} & \textbf{0.91}\\
\addlinespace[2pt]
\multicolumn{2}{@{}l}{\textit{Cross-round transition}}\\
Token projector & 0.02\\
Token-fusion PSG & 0.0001\\
\cmidrule(lr){1-2}
\textbf{Transition total} & \textbf{0.02}\\
\addlinespace[2pt]
\multicolumn{2}{@{}l}{\textit{Round 2: $240^2$, 6 heads, 384 channels}}\\
ViT path (without PSG or transition) & 5.34\\
Block PSG & 0.002\\
\cmidrule(lr){1-2}
\textbf{Round-2 stack total} & \textbf{5.34}\\
\addlinespace[2pt]
\textbf{Round-2 incremental total} & \textbf{5.36}\\
\textbf{Full two-round path} & \textbf{6.27}\\
\bottomrule
\end{tabular}
\caption{Compute decomposition of the default \name configuration. Counts follow the paper's GMACs convention and omit normalization, activations, interpolation, and elementwise operations.}
\label{tab:arch}
\end{table}

\section{Activation-Replacement Accuracy}
\label{sec:appendix-replacement-accuracy}
\begin{figure}[t]
\centering
\includegraphics[width=.8\columnwidth]{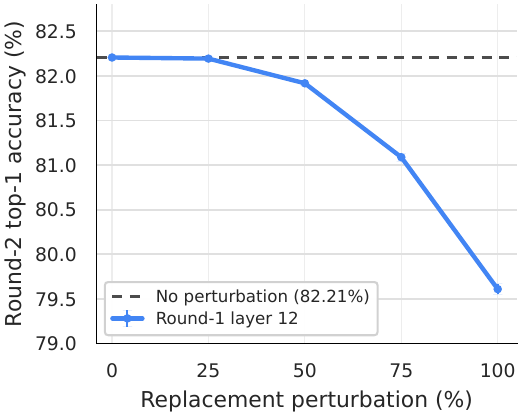}
\caption{Top-1 accuracy after replacing the round-1 block-12 output with activation-matched Gaussian noise.}
\label{fig:noise-accuracy}
\end{figure}

In this section, we further investigate the effect of round 1 on round 2 by replacing the reused round-1 activations with Gaussian noise matched to their mean and standard deviation. This controls for changes in activation magnitude and helps isolate the contribution of the information carried by the round-1 representation. Let $x$ denote the round-1 block-12 output. We define
\begin{equation}
x_{\alpha}=(1-\alpha)x+\alpha\epsilon,
\qquad \alpha\in[0,1],
\end{equation}
where $\epsilon$ is Gaussian noise with the scalar mean and standard deviation of the current activation tensor, and $\alpha$ controls the replacement strength. Figure~\ref{fig:noise-accuracy} reports top-1 accuracy over all 50,000 validation images at different replacement strengths. Accuracy decreases from 82.206\% without replacement to 79.610\% at full replacement, demonstrating that the information propagated from round 1 contributes to the accuracy of round 2.
 
\section{Activation-Replacement Confidence}
\label{sec:appendix-replacement-confidence}

\begin{figure}[t]
\centering
\includegraphics[width=.9\columnwidth]{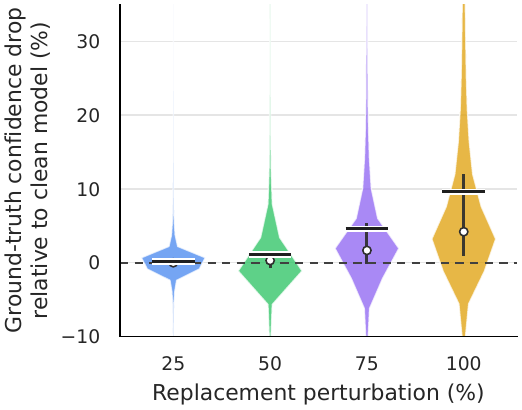}
\caption{Correct-class confidence under activation-matched replacement, evaluated on images classified correctly.}
\label{fig:noise-confidence}
\end{figure}

In addition to classification accuracy, the round-1 representation affects the confidence of round 2. Figure~\ref{fig:noise-confidence} reports correct-class confidence under the same activation-replacement intervention described in Appendix~\ref{sec:appendix-replacement-accuracy}. As the replacement strength increases, the confidence distribution shifts downward, indicating that removing information from the reused representation weakens the certainty of the round-2 predictions. This result complements the accuracy reduction and provides further evidence that round 2 meaningfully builds on the representation produced by round 1.

\begin{table}[t]
\centering
\setlength{\tabcolsep}{3.0pt}
\begin{tabular}{rccr}
\toprule
$\tau$ & GMACs & Top-1 (\%) & Exit distribution (\%)\\
& / image & & $128^2$ / $192^2$ / $240^2$\\
\midrule
0.00 & 7.097 & 82.130 & 0.0 / 0.0 / 100.0\\
0.10 & 6.246 & 82.130 & 1.4 / 14.1 / 84.5\\
0.20 & 5.353 & 82.126 & 6.0 / 24.8 / 69.2\\
0.30 & 4.814 & 82.110 & 10.2 / 29.4 / 60.4\\
0.40 & 4.414 & 82.038 & 13.8 / 32.3 / 53.9\\
0.50 & 4.074 & 81.940 & 17.2 / 34.3 / 48.6\\
0.60 & 3.765 & 81.822 & 20.5 / 35.7 / 43.8\\
0.70 & 3.472 & 81.608 & 23.8 / 37.0 / 39.2\\
0.80 & 3.158 & 81.226 & 27.2 / 38.5 / 34.4\\
0.90 & 2.868 & 80.614 & 30.9 / 39.0 / 30.1\\
1.00 & 2.597 & 79.886 & 34.6 / 39.3 / 26.1\\
1.20 & 2.105 & 78.208 & 42.5 / 38.4 / 19.2\\
1.40 & 1.667 & 75.814 & 51.2 / 35.3 / 13.5\\
1.60 & 1.254 & 73.002 & 60.7 / 30.8 / 8.5\\
1.80 & 0.872 & 69.186 & 71.4 / 24.2 / 4.5\\
2.00 & 0.526 & 64.442 & 83.6 / 14.8 / 1.6\\
2.20 & 0.242 & 58.970 & 96.8 / 3.2 / 0.1\\
2.30 & 0.188 & 57.626 & 100.0 / 0.0 / 0.0\\
\bottomrule
\end{tabular}
\caption{Entropy-routed operating points for the three-round $128^2\!\rightarrow\!192^2\!\rightarrow\!240^2$ model.}
\label{tab:three-round}
\end{table}

\begin{figure}
\centering
\includegraphics[width=.9\columnwidth]{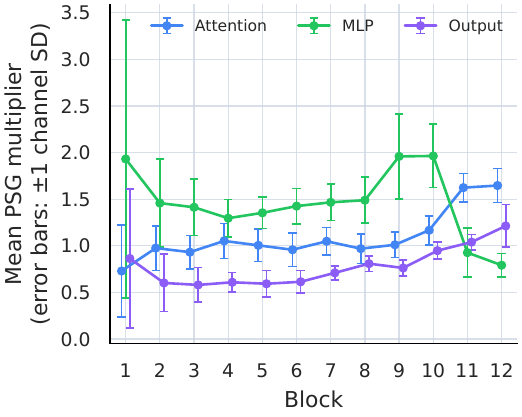}
\caption{Round-2 PSG means across depth. Error bars show the standard deviation across channels within each learned multiplier vector.}
\label{fig:psg-profile-summary}
\end{figure}

\section{Analytical Compute Decomposition}
\label{sec:appendix-compute}

Table~\ref{tab:arch} separates the cost of the shared ViT path, block-level PSG, and the cross-round transition. In round 1, the ViT path accounts for 0.91 of the 0.91 GMACs total, while block PSG adds 0.002 GMACs. The transition adds 0.02 GMACs, of which 0.02 comes from token projection and 0.0001 from the two fusion gates. The round-2 stack costs 5.34 GMACs, giving an incremental refinement cost of 5.36 GMACs and a cumulative two-round cost of 6.27 GMACs. Thus, the learned conditioning overhead is small relative to the transformer computation.

\section{PSG Specialization Across Depth}
\label{sec:appendix-psg-depth}

Figure~\ref{fig:psg-profile-summary} complements the channel-resolved PSG profiles in Figure~\ref{fig:psg-profile-values}. It summarizes the round-2 multipliers by branch and block. Each error bar is the standard deviation across active channels within one learned multiplier vector. Both the mean values and their channel-wise variation change across branches and depth, with particularly distinct modulation in the later blocks. These results show that PSG does not learn a single uniform rescaling; instead, it adjusts the attention, MLP, and block outputs differently according to their role in round-2 refinement.

\section{Three-Round Progression}
\label{sec:appendix-three-round}

To verify that \name is not restricted to two rounds, we train a three-round ImageNet-1K model progressing through resolutions of $128^2\!\rightarrow\!192^2\!\rightarrow\!240^2$ with 2, 4, and 6 heads, respectively. Executing all three rounds reaches 82.130\% top-1 accuracy at 7.097 GMACs. By varying the entropy threshold, this configuration provides operating points spanning 0.188--7.097 average GMACs and 57.626--82.130\% accuracy, as reported in Table~\ref{tab:three-round}. Although its peak accuracy is slightly below the 82.206\% achieved by the default two-round model, the additional exit point enables finer-grained routing across a broader compute range. Compared with the two-round version, this design is useful when input difficulty is highly heterogeneous, as easy samples can exit at substantially lower cost while more challenging samples receive additional computation.

\section{Attention Maps Across Progressive Rounds}
\label{sec:appendix-attention-maps}

\begin{figure*}
\centering
\includegraphics[width=1\textwidth]{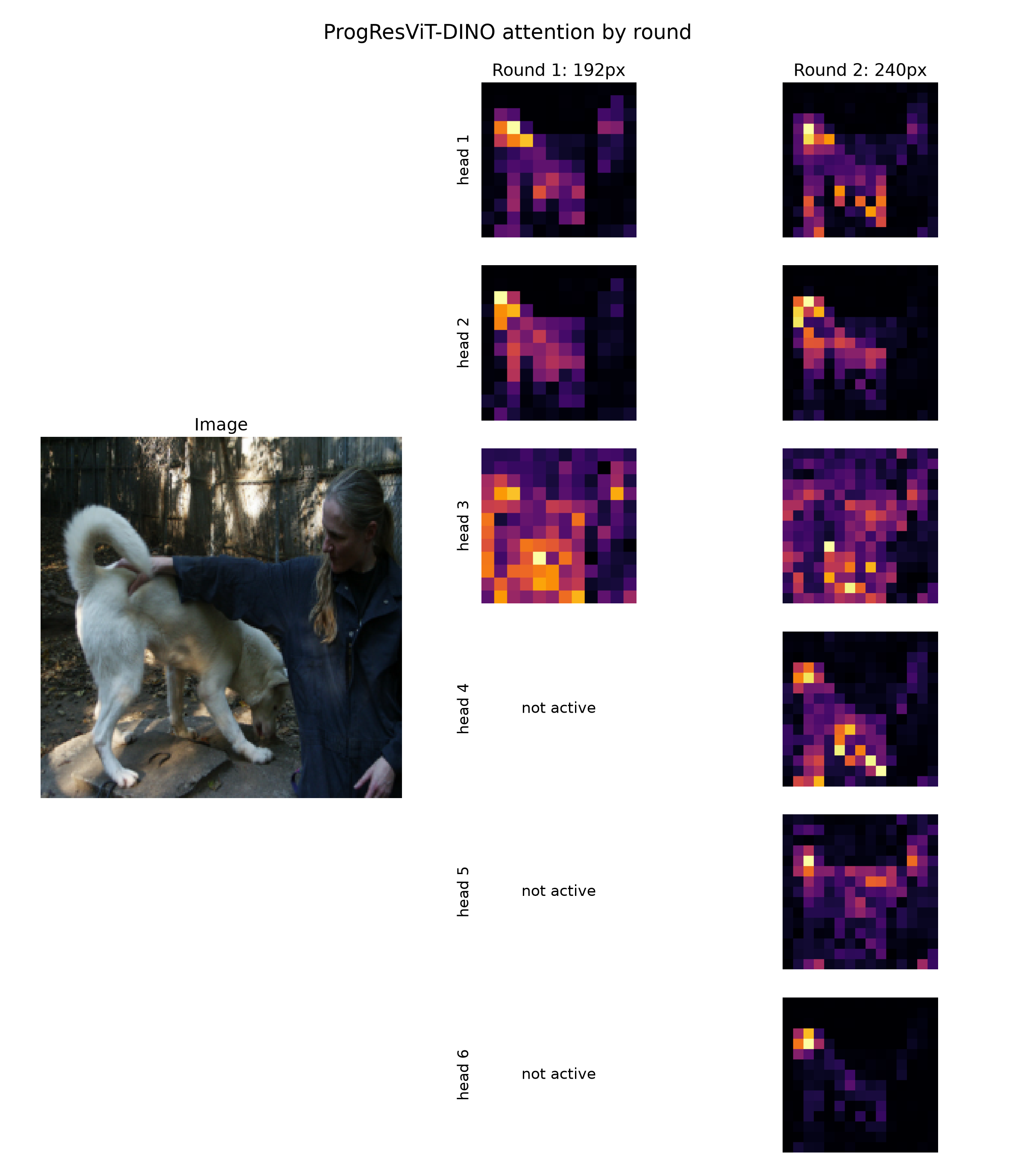}
\caption{Attention-map visualization for one ImageNet example from the Matryoshka DINO model. Round 1 uses the first three heads at low resolution. Round 2 retains these ordered head prefixes and activates heads 4--6 at higher resolution. Panels without an active round-1 head are marked as inactive.}
\label{fig:all-heads-rounds}
\end{figure*}

Figure~\ref{fig:all-heads-rounds} qualitatively visualizes the attention heads of \name implementation on DINO. Round 1 executes the first three heads at the lower resolution. At the higher resolution, round 2 reuses these shared prefix heads and activates three additional heads. The shared heads refine their round-1 attention patterns using the higher-resolution input, while heads 4--6 capture complementary information that differs from the patterns learned by the first three heads.

\end{document}